\documentclass[10pt,twocolumn,letterpaper]{article}

\usepackage[pagenumbers]{wacv} 

\usepackage{xcolor}
\usepackage{graphicx}
\usepackage{amsmath}
\usepackage{amssymb}
\usepackage{booktabs}
\usepackage{tabularx}

\usepackage[pagebackref,breaklinks,colorlinks]{hyperref}

\usepackage[capitalize]{cleveref}
\crefname{section}{Sec.}{Secs.}
\Crefname{section}{Section}{Sections}
\Crefname{table}{Table}{Tables}
\crefname{table}{Tab.}{Tabs.}

\def\wacvPaperID{59} 
\def\confName{WACV}
\def\confYear{2025}

\begin{document}

\title{Neural Bloom: A Deep Learning Approach to Real-Time Lighting}

\author{
    Rafa\l{} Karp\\
    {\tt\small karprafalkarp@gmail.com}
    \and
    Dawid Gruszka\\
    {\tt\small daw.grusz@gmail.com}
    \and
    Tomasz Trzci\'{n}ski\\
    IDEAS Research Institute\\
    Warsaw University of Technology\\
    Tooploox\\
    {\tt\small tomasz.trzcinski@pw.edu4.pl}
}

\maketitle

\begin{abstract}

We propose a novel method to generate bloom lighting effect in real time using neural networks. Our solution generate brightness mask from given 3D scene view up to 30\% faster 
than state-of-the-art methods. The existing traditional techniques rely on multiple blur appliances and texture sampling, also very often have existing conditional branching in its implementation. These operations occupy big portion of the execution time. We solve this problem by proposing two neural network-based bloom lighting methods, \textit{Neural Bloom Lighting (NBL)} and \textit{Fast Neural Bloom Lighting (FastNBL)}, focusing on their quality and performance. Both methods were tested on a variety of 3D scenes, with evaluations conducted on brightness mask accuracy and inference speed. The main contribution of this work is that both methods produce high-quality bloom effects while outperforming the standard state-of-the-art bloom implementation, with \textit{FastNBL} being faster by 28\% and \textit{NBL} faster by 12\%. These findings highlight that we can achieve realistic bloom lighting phenomena faster, moving us towards more realism in real-time environments in the future. This improvement saves computational resources, which is a major bottleneck in real-time rendering. Furthermore, it is crucial for sustaining immersion and ensuring smooth experiences in high FPS environments, while maintaining high-quality realism.
\end{abstract}

\section{Introduction}

Rendering in computer graphics is the process of generating an image from a model by simulating light, color, and materials to produce either realistic or stylized visual outputs. It can be broadly categorized into offline and real-time rendering. While offline rendering benefits from relaxed time constraints, achieving realism in real-time rendering is far more challenging due to computational limitations and the need for immediate results. A critical aspect of realism in both forms is light simulation, which replicates how light interacts with materials, surfaces, and the environment. Techniques such as reflection, refraction, and scattering are modeled to create realistic imagery that enhances visual appeal and immersion.

\begin{figure}
\begin{center}
\includegraphics[width=\linewidth]{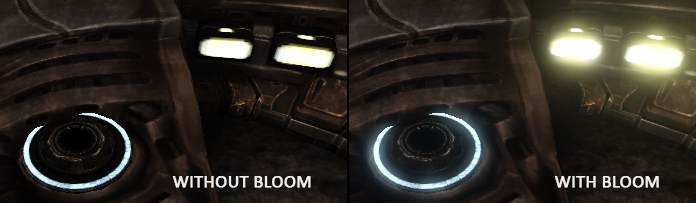}
\end{center}
\caption{Example of a bloom effect in OpenGL rendering \cite{LearnOpenGL2024}.}
\label{fig:bloom_example}
\end{figure}

One notable light simulation technique is the bloom effect, which simulates the phenomenon of bright light sources bleeding or glowing beyond their boundaries, replicating how cameras or human vision perceive intense lights \cite{Bavoil08, Kuster19, Pharr16}. Often implemented as a post-process effect using graphics APIs like OpenGL\cite{OpenGL2024}, bloom adds a soft halo around bright areas, enhancing the intensity of light sources in video games and cinematic, as illustrated in Figure ~\ref{fig:bloom_example}.

Realistic lighting is a fundamental challenge in real-time rendering due to the computational complexity involved in accurately simulating how light interacts with surfaces, materials, and the environment. Techniques like global illumination, soft shadows, and physically accurate reflections demand significant computational resources, which can strain performance, especially in applications requiring high frame rates. To address these challenges, the use of neural networks has emerged as a transformative solution, offering a way to approximate light behavior with remarkable speed and efficiency~\cite{MOHANDAS2019101499, Song2019NeuralIllumination}. Neural networks, trained on vast datasets of real-world lighting scenarios, dynamically estimate light sources, shadows, and reflections to mimic real-world lighting~\cite{lecun2015deep}. This capability not only alleviates the performance bottlenecks of traditional methods but also enables immersive graphics in gaming, augmented reality (AR), and virtual reality (VR). By balancing realism and computational efficiency, neural networks become core part of real-time rendering, enhancing user experiences and pushing the boundaries of what is possible in interactive applications.

In our research, we present a novel method for generating the Bloom lighting effect, Figure \ref{fig:bloom_example}], in real time using neural networks, achieving faster results than state-of-the-art deterministic implementations. The main motivation behind this work is to reduce the computational time required for realistic rendering on each frame, thereby freeing up processing resources that can be allocated to other effects, such as reflections, shadows, and physics-based interactions. This optimization aims to bring us a step closer to achieving greater realism in real-time rendering while also enabling people with less powerful hardware to experience improved visual quality and realism. We rigorously evaluate the performance and visual quality of our solution, comparing it with the implementation of Bloom lighting in the Unity3D game engine~\cite{Unity3D}, a popular standard in the industry, to demonstrate the advantages of our approach both in speed and rendering fidelity.

\subsection{Problem Statement}

In real-time rendering, achieving high frame rates (frames per second, or FPS) is essential for smoothness, responsiveness, and user immersion, particularly in interactive applications like gaming, simulations, and AR/VR. Frame rate targets such as 60 FPS (16.67 milliseconds per frame) or 120 FPS (8.33 milliseconds per frame) ensure fluid motion and minimal latency, critical for maintaining the user's experience. In VR, where rendering standards are even stricter, the minimum recommended frame rate is 90 FPS per eye~\cite{vr-90fps-standard}, requiring two simultaneous renders to prevent motion sickness~\cite{vr-motion-sickness} and ensure realism. High-end setups now aim for 240 FPS per eye, rendering frames in just 4.17 milliseconds, a standard that pushes the limits of even the most advanced rendering systems. This trend toward higher frame rates reflects the growing emphasis on ultra-responsive visuals, particularly in fast-paced environments, where smooth rendering is essential for user comfort and immersion.

As frame rate expectations rise, so do demands for realism in real-time rendering, creating new challenges for developers to meet performance and quality standards. Sophisticated lighting effects, such as bloom, reflections, and shadows, are critical for crafting believable environments, adding depth and richness to scenes. Users increasingly expect seamless integration of high frame rates with lifelike graphics, pushing rendering systems to simulate and render complex visual elements in real-time. Delivering this balance of high performance and visual fidelity remains a cornerstone of advancements in interactive rendering technologies, with continued innovation needed to meet these ever-growing demands.

In this research, we aim to address the performance challenges associated with a commonly used lighting effect—Bloom—by optimizing its speed without compromising visual quality. By achieving faster Bloom processing, our approach frees up valuable computational resources, enabling more accurate calculations for other effects within each frame. This improvement is intended to support increasingly realistic visual environments in real-time applications, meeting the demands of higher frame rates and enhanced graphical fidelity.

The main contributions of this work are as follows:
\begin{itemize}
  \item We propose a novel neural network-based method for generating the Bloom lighting effect in real time, optimized for low-latency applications.
  \item Our approach achieves faster performance compared to traditional Bloom implementations, such as those in Unity3D, while maintaining high visual quality.
  \item We provide a comprehensive evaluation that benchmarks both the visual fidelity and computational efficiency of our method against industry-standard techniques.
  \item Our results demonstrate the potential of neural rendering techniques to reduce rendering cost per frame, making advanced visual effects more accessible on lower-end hardware.
\end{itemize}

\section{Related Work}
\subsection{Bloom Effect in Real-Time Rendering}

One of the first implementations of real-time bloom effect can be found in Ico\cite{ico} video game game released on PlayStation 2 console, paving the way for later video games to incorporate bloom and other post-processing effects into their rendering pipelines. In 2004 the developers of game Tron 2.0 released an article \cite{glow-effect} detailing their implementation of then called "Real-Time Glow" effect, mainly designed for DirectX 8 vertex and pixel shader v1.1 with fallback implementation for DirectX 7 fixed-function model. This publication was crucial and started widespread adoption of the Bloom effect in the industry.

With new graphic APIs and programmable shaders introduced in mid-2000s, notably DirectX 9 with Shader Model 2.0 and OpenGL 2.0+, implementation techniques changed, shaders were used for high-pass filtering to separate bright areas into separate frame buffer and multi-scale Gaussian blur applied to it. In late-2000s, with DirectX 10 and OpenGL 3.0+ available, greater quality could be achieved, HDR rendering with 16bit per channel floating-point frame buffers allowed for greater accuracy and linear color space enabled more realistic light blending, as the GPU computational capacity increased additional mip-mapped pyramid blur was added to the algorithm greatly increasing visual fidelity. With the rise of Physically Based Rendering (PBR) and Post-Processing pipelines in 2010s, now possible thanks to DirectX 11, OpenGL 4+, early Vulkan versions and much faster GPUs, new techniques emerged, deferred rendering became standard in high-end rendering engines making post-process modular and highly parametrizable, using higher quality blur filters like Dual Filter Kawase and High-Quality Gaussian algorithms combined with logarithmic luminance and temporal filtering improved quality even more. Today with modern Graphics APIs like DirectX 12, Vulkan and Metal available, implementations are slowly moving into more flexible and efficient compute shaders with more emphasis on more realism, introducing lens flares, dirt textures, chromatic aberration and more.

\subsection{Machine Learning for Real-Time Rendering}

Machine Learning is becoming increasingly integrated into real-time rendering, addressing problems of lighting, shadowing, under-sampling, post-processing effects and more.
CNN for Screen-Space Shading \cite{nalbach2016deep} is an example of a network striving to combine previously separate screen-space lighting effects into one, effectively simplifying the rendering pipeline.
Neural Supersampling \cite{neural-super-sampling} demonstrates that CNNs are a viable solution for real-time upsampling of images when provided with additional depth and dense motion vectors data for temporal consistency.

Neural Network Ambient Occlusion (NNAO) \cite{nnao} shows that a shallow network can generate an accurate representation of a commonly used Screen Space Ambient Occlusion effect while also addressing visual artifacts created by shader implementation.

\section{Method}

The traditional bloom effect, as analyzed in detail, involves extracting bright regions of a scene and applying a Gaussian blur~\cite{haas2001digital}, which requires multiple passes of texture sampling at varying resolutions. This process can be computationally expensive for several reasons. First, the extraction of bright regions relies on a thresholding step that introduces conditional logic. On GPUs, such conditional branching can lead to warp divergence, where some pixels follow different execution paths, reducing parallel efficiency. Second, the Gaussian blur itself involves sampling multiple neighboring texels (\textit{texture elements}) per pixel in each pass, with each sample requiring memory access and interpolation. A full Gaussian blur usually requires separate horizontal and vertical passes, multiplying the number of texture samples required. Third, the use of multiple passes inherently adds overhead, as each pass involves rendering the output into an intermediate texture and reading it back for subsequent operations, which incurs additional memory bandwidth and latency costs.

The above-described issues can be mitigated by using a neural network. Unlike traditional methods, a neural network does not require multiple texture sampling or separate conditioning to approximate the blur effect. Additionally, it can inherently extract and process the necessary threshold values within its learned parameters. Our analysis uncovered that a simple spatially aware operation can encapsulate all the necessary information and generate brightness mask texture with significantly fewer operations than traditional bloom. Neural network-based approach eliminates the need for iterative texture sampling and conditional branching by directly generating the desired glow effect in a single pass, potentially offering a more efficient alternative.

\begin{figure}
\begin{center}
\includegraphics[width=0.8\linewidth]{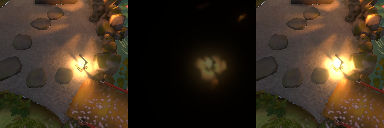}
\end{center}
\caption{\textit{Left:} Selected view of the 3D scene without bloom. \textit{Middle:} Isolated bloom glow effect (brightness mask). \textit{Right:} Same view with the brightness mask applied. Each image is 128×128 pixels.}
\label{fig:bloom_mask_addition_example}
\end{figure}

We began by exploring various neural network architectures for generating the bloom effect. 
Given the importance of spatial features in the bloom effect an architecture that could capture local patterns and structures in the image was needed, that lead us to focus on convolutional networks and ultimately resulted in U-Net-inspired architecture~\cite{ronneberger2015u}, which provides a flexible framework for efficient feature extraction.
The U-Net design, with its encoder-decoder structure, facilitates the extraction of various spatial features at different levels of abstraction. This capability made it an ideal choice for our task, as it allowed us to capture and process the necessary image features for generating the bloom effect, all while maintaining efficient computation. Moreover, the shallow nature of the architecture ensured that the network remained lightweight and quick to process, striking a balance between performance and speed. The input of our neural network was an image containing concrete view on the 3D scene, the output was a brightness mask texture, highlighting bright scene regions. By adding the input image with generated bloom glow effect, we are able to achieve illuminated view on given 3D scene (as illustrated in Figure \ref{fig:bloom_mask_addition_example}). The generated bloom effect can be seamlessly integrated with the original scene using a shader, resulting in a final composite image. For this study, we chose to work with 128×128 pixel images. Moreover, to enhance the fidelity of the generated results, we trained each neural network model exclusively on a single, selected 3D scene

We trained our solutions on machine with single NVIDIA L4 GPU. To select hyper-parameters we conducted extensive research and series of experiments. The models were trained on 32 image batch size. We used \textit{Adam optimizer} with static \textit{learning rate} equal to 0.0002. We experimented with multiple different loss functions, but in the end we used \textit{MSELoss}~\cite{pytorch_mse_loss} (PyTorch implementation). We trained our methods on 1500 epochs. Loss was calculated on difference between generated Bloom glow effect vs. expected Bloom glow effect. In the final stage of our experiments, we identified two distinct solutions: \textit{Neural Bloom Lighting (NBL)}, which provides superior quality at the cost of slower performance, and \textit{Fast Neural Bloom Lighting (FastNBL)}, which is faster but results in reduced quality. These solutions will be referred to as \textit{NBL} and \textit{FastNBL} respectively for the remainder of this paper. Below, we provide a detailed description of each solution, focusing on the design and implementation of the respective neural networks. Below, we provide a detailed description of each solution, focusing on the design and implementation of the respective neural networks. It is important to note that these two neural networks do not differ fundamentally in their architectural design, but rather in their parametric configurations and optimization strategies.

\begin{figure*}
\begin{center}
\includegraphics[width=0.99\textwidth]{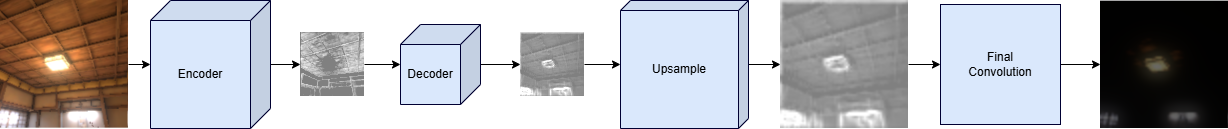}
\end{center}
\caption{\textit{Neural Bloom Lighting (NBL)} architecture visualized.}
\label{fig:nbl-architecture}
\end{figure*}

{\textit{Neural Bloom Lighting (NBL)}} architecture consists of the following components (visualized in Figure \ref{fig:nbl-architecture}):
    \begin{enumerate}
        \item \textbf{Encoder}: A convolutional block (\texttt{ConvBatchRelu}) that processes the input (\(N, 3, H, W\)) to produce feature maps of size (\(N, 64, H/2, W/2\)). It includes: a convolutional layer (\texttt{nn.Conv2d}) with a kernel size of 3, stride of 2, and 64 output channels, batch normalization (\texttt{nn.BatchNorm2d}), and the ReLU activation function (\texttt{nn.ReLU}).
        \item \textbf{Decoder}: A second convolutional block (\texttt{ConvBatchRelu}) that reduces the feature dimensions from 64 to 32 channels, maintaining the resolution (\(H/2, W/2\)).
        \item \textbf{Upsampling}: Bilinear interpolation (\texttt{nn.Upsample}) scales the feature maps back to the original spatial dimensions (\(H, W\)).
        \item \textbf{Output}: A final convolutional layer (\texttt{nn.Conv2d}) with a kernel size of 1, reducing the 32 channels to 3 output channels, followed by a HardTanh activation (\texttt{nn.Hardtanh}) to constrain the output range to \([-1, 1]\).
    \end{enumerate}

\textit{Fast Neural Bloom Lighting (FastNBL)} architecture consists of the following components:
    \begin{enumerate}
        \item \textbf{Encoder}: A dilated convolutional block (\texttt{ConvBatchReluDilated}) that processes the input (\(N, 3, H, W\)) to produce feature maps of size (\(N, 32, H/2, W/2\)). It includes: A dilated convolutional layer (\texttt{nn.Conv2d}) with a kernel size of 3, stride of 2, dilation of 2, and 32 output channels, batch normalization (\texttt{nn.BatchNorm2d}) and ReLU activation function (\texttt{nn.ReLU}).
        \item \textbf{Decoder}: Another dilated convolutional block (\texttt{ConvBatchReluDilated}) with grouped convolutions (\texttt{groups=32}), maintaining the same spatial dimensions (\(H/2, W/2\)) and outputting 32 channels.
        \item \textbf{Upsampling}: Bilinear interpolation (\texttt{nn.Upsample}) scales the feature maps from (\(H/2, W/2\)) to the original input dimensions (\(H, W\)).
        \item \textbf{Output}: A final convolutional layer (\texttt{nn.Conv2d}) with a kernel size of 1, reducing the 32 channels to 3 output channels. This is followed by a HardTanh activation (\texttt{nn.Hardtanh}) to constrain the output range to \([-1, 1]\).
    \end{enumerate}

\section{Results}

The results presented in this section focus on the two methods introduced in the previous chapter: \textit{Neural Bloom Lighting (NBL)} and \textit{Fast Neural Bloom Lighting (FastNBL)}. While the primary evaluations were conducted on a single 3D scene to maintain consistency, additional training and testing were performed on a variety of other scenes to demonstrate the flexibility and robustness of the proposed methods. The process of acquiring these scenes, along with the synthetic preparation of generated data used for training and evaluation, is described in Section~\ref{sec:data_generation}.

Our evaluations were centered on two key aspects: \textbf{quality} and \textbf{performance}. The quality assessment measured how accurately the generated brightness masks replicated the desired illumination effects, while the performance evaluation analyzed the inference speed of both methods. These assessments are detailed in the following subsections: \textit{Quality Evaluation} ~\ref{sec:quality_evaluation} and \textit{Generation Performance Assessment} ~\ref{sec:perf_evaluation}. Finally, a comparison with state-of-the-art bloom techniques is provided to contextualize the results and is further discussed in the subsequent section (\ref{sec:comparison_with_state_of_the_art_bloom}).

\subsection{Data Generation} \label{sec:data_generation}

\begin{figure}
\begin{center}
    \includegraphics[width=0.49\linewidth]{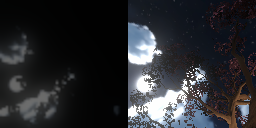} \hfill
    \includegraphics[width=0.49\linewidth]{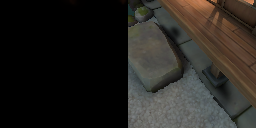}
    
    \vspace{0.3em} 
    
    \includegraphics[width=0.49\linewidth]{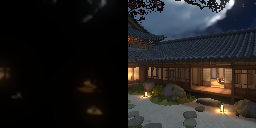} \hfill
    \includegraphics[width=0.49\linewidth]{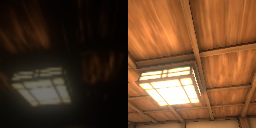}
\end{center}
\caption{Example of four sample input data pairs. The right column shows captured viewports from 3D scene, while the left column displays corresponding brightness maps generated by the Unity3D shaders}
\label{fig:sample-of-input-images}
\end{figure}

For our study, we prepare and synthetically generate the training and evaluation data. Using the Unity Asset Store~\cite{unity_asset_store}, we prepared 15 3D scenes featuring diverse elements and various light sources. Each scene was automatically navigated, and camera viewports were captured at a resolution of 128x128 pixels, this resolution was chosen to maximize iteration speed of this project. Subsequently, Unity3D's bloom shader was applied to these viewports to generate brightness masks, representing the bloom glow effect.

Rendering is done in Unity 3D v2022.3.48f using the Universal Rendering Pipeline (URP) v14.0.11 with Bloom Effect \cite{Unity3D-URP-Bloom} parameters as provided in Table ~\ref{tab:bloom_parameters}, configured on global volume with a weight of 1.0.
\begin{table}[h!]
\centering
\begin{tabular}{ |c|c| } 
 \hline
 Threshold & 0.9 \\ \hline
 Intensity & 1 \\ \hline
 Scatter & 0.5 \\ \hline
 Tint & (1,1,1) \\ \hline
 Clamp & 65472 \\ \hline
 High Quality Filtering & Enabled \\ \hline
 Downscale & Half \\ \hline
 Max Iterations & 8 \\ \hline
 Dirt Texture & Disabled \\ \hline
\end{tabular}
\caption{Unity3D Bloom Post Processing effect parametrs.}
\label{tab:bloom_parameters}
\end{table}

Finally, the captured viewport and its corresponding brightness mask were combined into a single concatenated image. Examples of these paired images are shown in Figure~\ref{fig:sample-of-input-images}.

For each 3D scene, we prepared a test dataset comprising 5,000 paired images, which were subsequently used in our evaluations (as detailed in the following sections).

\subsection{Quality Evaluation} \label{sec:quality_evaluation}

We determined that the most effective metric for comparing the generated bloom brightness masks to the intended ground truth is the \textbf{Pixel Mean Squared Error (MSE)}. For our implementation, we utilized the PyTorch library to compute this metric. The evaluation was conducted on the entire test dataset (as described in Section~\ref{sec:data_generation}), focusing on two key values: the \textbf{average MSE} and the \textbf{99th percentile (p99) MSE}.

Through experimentation, we observed that MSE values below \(0.001\) are nearly indistinguishable from the ground truth images. Consequently, our target was to achieve an average MSE lower than \(0.001\), aiming for values as close as possible to \(0.0001\). 

For the \(p99\) MSE, we manually inspected the worst-performing percentile of images to analyze discrepancies between the generated and expected outputs. This analysis revealed that higher MSE values predominantly occurred in images with higher brightness levels. Interestingly, for these images, because they are so bright, the large differences in brightness masks become imperceptible when combined with the input images (3D scene viewports). As a result, the differences in the final viewport images with bloom effects (generated by our methods or the Unity3D shader) are not easily noticeable.

Further comparisons and visual examples illustrating these findings are provided in Section~\ref{sec:comparison_with_state_of_the_art_bloom}.

\subsection{Generation Performance Evaluation} \label{sec:perf_evaluation}

As our research focuses on accelerating bloom generation, it was essential to establish a reliable performance comparison between our proposed solutions and a traditional state-of-the-art shader-based method. We measured the generation times of bloom brightness masks using a standard shader-based Bloom implementation and our models (\textit{NBL} and \textit{FastNBL}). For each measurement, we excluded the top 2\% of the slowest and fastest times to ensure a robust comparison. Additionally, each image was processed 50 times, and the average of these times was used. All time measurements were performed using CUDA~\cite{nvidia2023cuda} to capture GPU-invoked calculations.

\begin{figure*}
\begin{center}
\includegraphics[width=0.99\textwidth]{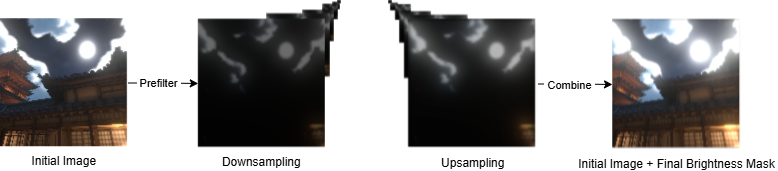}
\end{center}
\caption{\textit{Shader Based Bloom Effect} passes visualized.}
\label{fig:bloom-passes}
\end{figure*}

We implement the shader based solution in Python using moderngl package\cite{modern-gl-package} v5.8.0, widely used OpenGL API wrapper, performance statistics are captured using Query Objects~\cite{open-gl-query-objects} with target parameter set to GL\_TIME\_ELAPSED, all shaders are compiled with \#version 430 core. Our bloom algorithm is based on the commonly used 3 phase solution, visualized in Figure \ref{fig:bloom-passes}, generally accepted as the standard way to implement the robust bloom effect, among others, used by the Unity3D Engine URP renderer \cite{Unity3D-URP-Bloom}. Main work consists of 3 passes:
\begin{enumerate}
    \item \textbf{Prefilter}: extracting parts of image where pixel brightness is above given threshold value into separate frame buffer.
    \item \textbf{Downsampling}: resizing prefiltered frame-buffer sequentially, halving its resolution in each iteration and applying 2 phase Gaussain blur, until the minimum mipmap size is achieved.
    \item \textbf{Upsampling}: combining all previously generated mipmaps backward using bicubic sampling, into a single final brightness mask.
\end{enumerate}
This directly reflects Unity3D URP renderer bloom effect algorithm, all shader parameters drive the same characteristics and are set to the same values used for generating training data in Unity3D Engine. \ref{tab:bloom_parameters}
Since all operations are applied sequentially, frame buffers generated during downsampling can be reused during upsampling pass, and the final brightness mask is saved into frame buffer holding prefiltered texture.

For our proposed method, performance evaluations were conducted for both trained models (\textit{NBL} and \textit{FastNBL}) using the same approach. After loading the weights of the prepared model, we fused the Convolution, BatchNormalization, and ReLU activation functions for optimized inference. The model was then converted into the TorchScript format~\cite{pytorch2019torchscript}. Finally, we executed the model inference for each input in the test dataset and measured the time taken for bloom brightness map generation.

\subsection{Final Comparison} \label{sec:comparison_with_state_of_the_art_bloom}

In this section, we present a comprehensive performance and visual quality comparison between our proposed methods (\textit{Neural Bloom Lighting (NBL)} and \textit{Fast Neural Bloom Lighting (FastNBL)}) and the state-of-the-art Unity3D bloom. Building upon the comparison methodology described earlier, we evaluate the methods across key metrics, including computational efficiency and output quality. 

\begin{figure*}
\begin{center}
\includegraphics[width=0.99\textwidth]{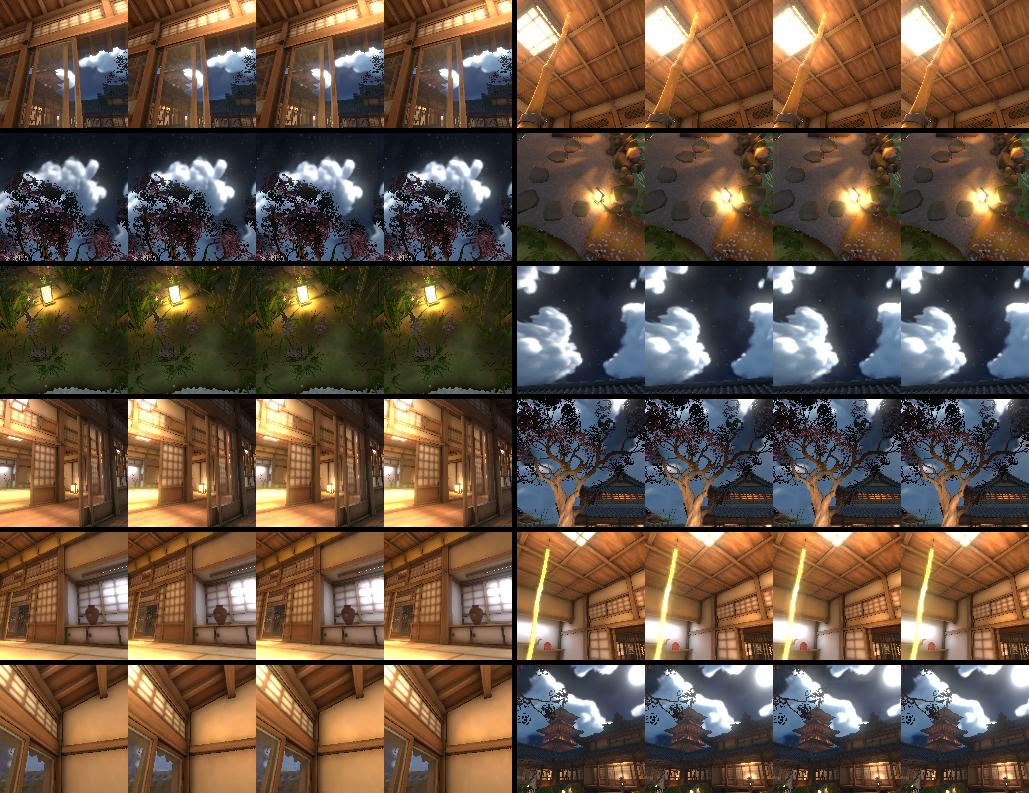}
\end{center}
\caption{The figure contains 12 examples arranged in 2 columns. \textit{Left:} Top-left view of the 3D scene without bloom. \textit{Middle Left:} Same viewport with bloom generated using \textit{Neural Bloom Lighting (NBL)}. \textit{Middle Right:} Bloom generated using  \textit{Fast Neural Bloom Lighting (FastNBL)}. \textit{Right:} Bloom generated using Unity3D's built-in bloom effect. Each image is 128×128 pixels.}
\label{fig:generated_images_example1}
\end{figure*}

Below, we provide visual examples (see Figure~\ref{fig:generated_images_example1}) showcasing the final results, which include the rendered 3D viewport combined with the generated brightness mask, using our two techniques, \textit{Fast Bloom Lighting (FastNBL)} and \textit{Neural Bloom Lighting (NBL)}, alongside Unity3D's bloom implementation. These images were selected randomly from a pool of 5000 examples, and the average MSE for these images (combined for both methods) is \textit{0.0008}.

During our testing (described in section~\ref{sec:quality_evaluation}), conducted on a set of 5000 images, the average MSE for \textit{Neural Bloom Lighting (NBL)} was \textit{0.00029}, while for \textit{Fast Bloom Lighting (FastNBL)} it was \textit{0.00076}. Looking at these images and comparing them to our average MSE, we can observe that the quality of these images is very high, with visual differences being indistinguishable.

\begin{figure*}
\begin{center}
\includegraphics[width=0.99\textwidth]{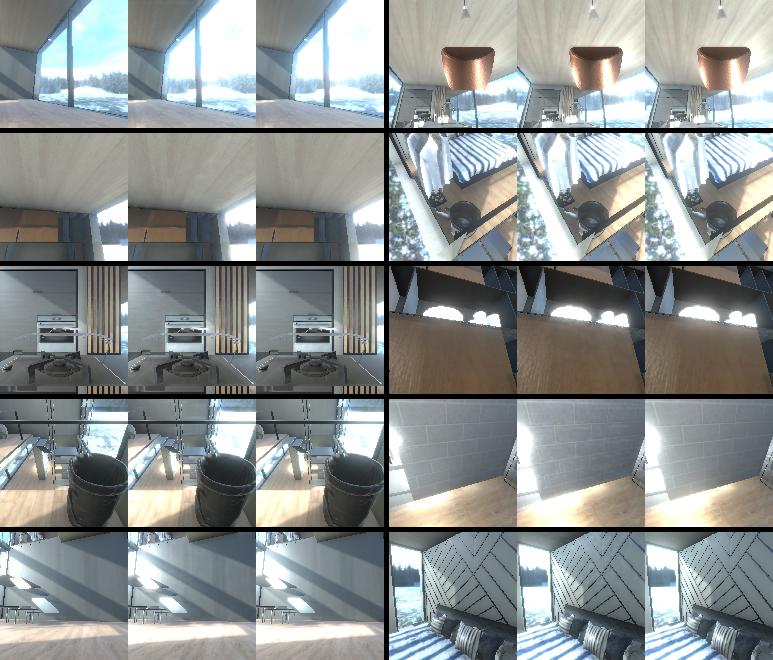}
\end{center}
\caption{The figure contains 10 examples arranged in 2 columns. \textit{Left:} Top-left view of the 3D scene without bloom. \textit{Middle:} Same viewport with bloom generated using \textit{Neural Bloom Lighting (NBL)}.  \textit{Right:} Bloom generated using Unity3D's built-in bloom effect. Each image is 128×128 pixels. The images were from different 3D scene that other figures in this scientific paper.}
\label{fig:generated_images_example2_nbl_other_scene}
\end{figure*}

We also applied the same method to a different scene. Below, in Figure~\ref{fig:generated_images_example2_nbl_other_scene}, we present the visually rendered results using the same \textit{Neural Bloom Lighting (NBL)} method for this other scene.

To evaluate the performance of the various bloom techniques, we measured their execution times in milliseconds across several metrics, including percentiles (p01, p05, p25, p50, p75, p95, p99) and the average. These metrics provide insights into the behavior of each method under varying conditions, from the fastest execution times (p01) to the slowest (p99). The measurements for \textit{Fast Neural Bloom Lighting (FastNBL)}, \textit{Neural Bloom Lighting (NBL)}, and Unity3D Bloom are presented in Table~\ref{tab:bloom_times}. These values offer a comparative view of performance, helping to assess both efficiency and consistency among the methods.

\begin{table*}[ht!]
\centering
\resizebox{0.99\textwidth}{!}{%
\begin{tabularx}{\textwidth}{|l|>{\raggedleft\arraybackslash}X|>{\raggedleft\arraybackslash}X|>{\raggedleft\arraybackslash}X|}
\hline
\textbf{Metric} & \textbf{\textit{Fast Neural Bloom Lighting}} & \textbf{\textit{Neural Bloom Lighting}} & \textbf{Unity3D Bloom} \\ \hline
p01             & 0.11677                & 0.13277                    & 0.13610                \\ \hline
p05             & 0.11741                & 0.13387                    & 0.16129                \\ \hline
p25             & 0.12054                & 0.13735                    & 0.14319                \\ \hline
p50             & 0.12361                & 0.14011                    & 0.17317                \\ \hline
p75             & 0.12546                & 0.14268                    & 0.17555                \\ \hline
p95             & 0.13017                & 0.14882                    & 0.17831                \\ \hline
p99             & 0.13985                & 0.15896                    & 0.18020                \\ \hline
average         & 0.12352                & 0.14053                    & 0.17253                \\ \hline
\end{tabularx}%
}
\caption{Performance comparison of \textit{Fast Neural Bloom Lighting (FastNBL)}, \textit{Neural Bloom Lighting (NBL)}, and Unity3D Bloom across different metrics. Times are in milliseconds (ms).}
\label{tab:bloom_times}
\end{table*}

From Table~\ref{tab:bloom_times}, several observations can be made. On average, \textit{FastNBL} performs approximately \textbf{28.4\% faster} than Unity3D Bloom (\textit{0.12352 ms vs. 0.17253 ms}) and about \textbf{12.1\% faster} than \textit{NBL} (\textit{0.12352 ms vs. 0.14053 ms}). When considering the $99^{\text{th}}$ percentile (p99), \textit{FastNBL} still outperforms the other methods, with a runtime that is \textbf{22.4\% lower} than Unity3D Bloom and \textbf{12.0\% lower} than \textit{NBL}. However, \textit{NBL} shows more consistent performance across percentiles compared to Unity3D Bloom, which exhibits a larger variation between its fastest and slowest times. These results highlight the trade-offs between speed and quality, with \textit{FastNBL} excelling in efficiency, while \textit{NBL} provides a balanced alternative, and Unity3D Bloom demonstrates less predictable performance.

The results presented above demonstrate that our bloom methods achieve a level of quality indistinguishable from state-of-the-art implementations. Despite this, our methods are significantly faster, with \textit{Fast Neural Bloom Lighting (FastNBL)} outperforming both \textit{Neural Bloom Lighting (NBL)} and Unity3D Bloom across all percentiles and the average. These findings highlight the efficiency of our approaches, which provide competitive performance without sacrificing quality, making them a superior choice for real-time applications.

\section{Conclusions and Future Work}

We developed a solution to generate a bloom lighting effect more efficiently than current state-of-the-art shader implementations. As our results demonstrate, we achieve a high-quality bloom brightness mask in less average time compared to traditional methods. This represents a significant step toward enhancing realism in real-time rendering environments and simulations. By employing a neural network-based single-pass approach, our method reduces the computational overhead in real-time graphics pipelines while maintaining a high level of realism in light simulation. Moreover, this work highlights the potential of neural networks to be applied in a wide range of rendering scenarios, opening the door for further innovations aimed at decreasing computational load. Future research could extend these principles to optimize other aspects of real-time graphics, such as shading, reflections, or complex lighting effects.

While the proposed approach demonstrates significant results, certain limitations need to be addressed in future work. First, to enable direct integration into various game engines or SDKs, the solution requires compatibility with a wide range of model inference performance optimizations. Without such enhancements, generating the brightness mask may not achieve faster performance compared to traditional solutions. Second, the current models were trained on a specific 3D scene, limiting their adaptability. To reduce resource demands and increase practical applicability, future models should be more flexible, capable of predicting brightness masks across diverse scenes. Finally, the proposed solution should support variable input and output resolutions. Most real-time environments operate at resolutions much higher than 128×128, so extending the method to handle larger resolutions is essential for broader adoption.

Future research could focus on addressing the limitations highlighted in the previous paragraph, ensuring broader applicability and robustness of the proposed approach across diverse rendering scenarios. In addition, extending similar neural network-based techniques to simulate other light phenomena—such as reflections, refractions, and scattering—offers an exciting direction. Beyond lighting, this methodology could be adapted to optimize computationally intensive tasks in real-time rendering, such as physics-based simulations, texture generation, or procedural content creation. These advancements could further reduce computational loads while maintaining the visual fidelity required for immersive environments and high-performance applications.


{\small
\bibliographystyle{ieee_fullname}
\bibliography{egbib.bib}
}

\end{document}